\pdfoutput=1
\documentclass{article}

\PassOptionsToPackage{numbers, compress}{natbib}

\usepackage[main, final]{neurips_2026}

\usepackage[utf8]{inputenc} 
\usepackage[T1]{fontenc}    
\usepackage{hyperref}       
\usepackage{url}            
\usepackage{booktabs}       
\usepackage{multirow}       
\usepackage{graphicx}       
\usepackage{amsfonts}       
\usepackage{nicefrac}       
\usepackage{microtype}      
\usepackage{xcolor}         

\newcommand{\zcomment}[1]{}
\newcommand{\zedit}[1]{#1}

\newcommand{\cready}[1]{#1}

\newcommand{\crebut}[1]{#1}
\newenvironment{crebutsec}{}{}

\usepackage{amsmath}
\usepackage{pifont}

\title{AVO: Agentic Variation Operators for \\Autonomous Evolutionary Search}

\author{%
  Terry Chen\thanks{Equal Contribution}~, Zhifan Ye$^*$, Bing Xu$^*$, Zihao Ye, Timmy Liu,  Ali Hassani, Tianqi Chen\\
\textbf{Andrew Kerr, Haicheng Wu, Vedaanta Agarwalla, Fengzhe Zhou, Harry Li, Yang Xu}\\
  \textbf{Yu-Jung Chen, Hanfeng Chen, Aditya Kane, Ronny Krashinsky, Ming-Yu Liu} \\
  \textbf{Vinod Grover, Luis Ceze, Roger Bringmann, John Tran, Wei Liu}\\
  \textbf{Fung Xie, Michael Lightstone, Humphrey Shi}\\
  NVIDIA\\
}

\begin{document}

\maketitle

\begin{abstract}
Agentic Variation Operators (AVO) are a new family of evolutionary variation operators that replace the fixed mutation, crossover, and hand-designed heuristics of classical evolutionary search with autonomous coding agents. 
Rather than confining a language model to \textbf{candidate generation within a prescribed pipeline}, AVO instantiates \textbf{variation as a self-directed agent loop} that can consult the current lineage, a domain-specific knowledge base, and execution feedback to propose, repair, critique, and verify implementation edits.
We evaluate AVO on \textbf{attention}, among the most aggressively optimized kernel targets in AI, on NVIDIA Blackwell (B200) GPUs. Over 7 days of continuous autonomous evolution on multi-head attention, AVO discovers kernels that outperform \textbf{cuDNN by up to 3.5\%} and \textbf{FlashAttention-4 by up to 10.5\%} across the evaluated configurations. The discovered optimizations transfer readily to grouped-query attention, requiring only \textbf{30 minutes} of additional autonomous adaptation and yielding gains of up to \textbf{7.0\% over cuDNN} and \textbf{9.3\% over FlashAttention-4}. 
Together, these results show that agentic variation operators move beyond prior LLM-in-the-loop evolutionary pipelines by elevating the agent from candidate generator to variation operator, and can discover performance-critical micro-architectural optimizations that produce kernels surpassing state-of-the-art expert-engineered attention implementations on today’s most advanced GPU hardware.
\end{abstract}

\section{Introduction}
\label{sec:introduction}

Large language models have emerged as powerful components in evolutionary search, replacing hand-crafted mutation operators~\citep{o2010open} with learned code generation~\citep{lehman2022elm, funsearch2024, alphaevolve2025, chen2023evoprompting}.
In these systems, an LLM generates candidate solutions conditioned on selected parents, while a surrounding framework, which is usually heuristic-based, handles parent sampling, evaluation, and population management.
This combination has produced notable results in mathematical optimization and algorithm discovery, including flagship systems such as FunSearch and AlphaEvolve~\citep{funsearch2024, alphaevolve2025}.
However, confining the LLM to candidate generation within a prescribed pipeline fundamentally limits what the LLM can discover: it produces a single output per invocation, with no ability to proactively consult reference materials, test its changes, interpret feedback, or revise its approach before committing a candidate.
For the most aggressively hand-tuned implementations, where further improvement requires deep, iterative engineering, this constraint is especially limiting.

We study this problem in the context of attention~\citep{vaswani2017attention}, the central operation in Transformer architectures, and one of the most heavily optimized GPU kernels.
The FlashAttention lineage~\citep{dao2022flashattention, dao2024flashattention2, shah2024flashattention3, zadouri2026flashattention4} and NVIDIA's cuDNN library~\citep{chetlur2014cudnn} have pushed attention throughput progressively closer to hardware limits across successive GPU generations, with both FlashAttention-4 (FA4) and cuDNN requiring months of manual optimization on the latest Blackwell architecture.
Surpassing these implementations demands sustained, iterative interaction with the development environment: studying hardware documentation, analyzing profiler output to identify bottlenecks, implementing and testing candidate optimizations, diagnosing correctness failures, and revising strategy based on accumulated experience.

\begin{figure}[t]
  \centering
    \includegraphics[width=0.9\textwidth]{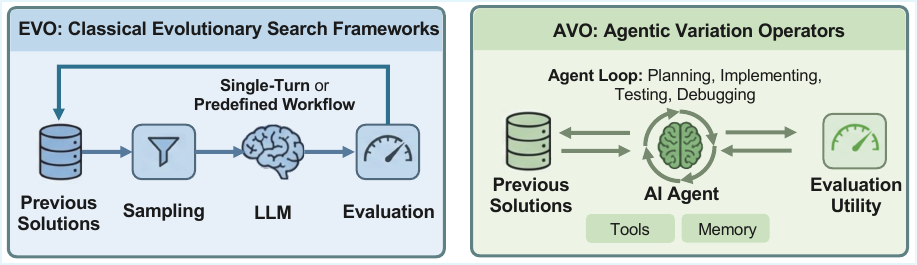}
  \caption{\textbf{EVO vs AVO}: Comparison between prior evolutionary search frameworks (e.g. FunSearch, AlphaEvolve, and related LLM-augmented evolutionary approaches) and the proposed Agentic Variation Operator. \textbf{Left:} Prior approaches follow a fixed pipeline where the LLM is confined to a single-turn generation step or a predefined workflow, with sampling and evaluation controlled by the framework. \textbf{Right:} AVO replaces this pipeline with an autonomous AI agent that iteratively plans, implements, tests, and debugs across long-running sessions, with direct access to previous solutions, evaluation utilities, tools, and persistent memory.}
  \label{fig:teaser}
\end{figure}

Recent progress in \emph{deep agents}~\citep{jimenez2024swebench, yang2024sweagent, wang2024openhands, anthropic2025cc, openai2025codex} demonstrates that LLMs augmented with planning, persistent memory, and tool use can autonomously navigate such multi-step engineering workflows, with applications ranging from resolving complex GitHub issues to generating key deep learning software~\citep{xu2026vibetensor}.
This motivates a fundamentally different role for LLMs in evolutionary search: rather than confining them within a fixed pipeline, we can elevate a deep agent to serve as the variation operator itself.
To this end, we propose \textbf{Agentic Variation Operators (AVO)}, in which a self-directed coding agent replaces the mutation and crossover process in previous works based on single-turn LLMs~\citep{funsearch2024, alphaevolve2025, chen2023evoprompting} or fixed workflows~\citep{wan2025loongflow}.
The AVO agent has access to all prior solutions, a domain-specific knowledge base, and the evaluation utility.
It autonomously decides what to consult, what to edit, and when to evaluate, enabling continuous improvements over extended time horizons.

To demonstrate its effectiveness, we apply AVO to multi-head attention (MHA) kernels on the Blackwell B200 GPU, and directly compare against the expert-optimized cuDNN and FlashAttention-4 kernels.
Over 7 days of continuous evolution without human intervention, the agent explored over 500 optimization directions and evolved 40 kernel versions, producing MHA kernels achieving up to 1668 TFLOPS at BF16 precision, outperforming cuDNN by up to 3.5\% and FlashAttention-4 by up to 10.5\%.
Our analysis of agent-discovered optimizations reveals that they span multiple levels of kernel design, including register allocation, instruction pipeline scheduling, and workload distribution, reflecting genuine hardware-level reasoning.
Empirically, we find that the optimization techniques discovered on MHA transfer effectively to grouped-query attention (GQA): adapting the evolved MHA kernel to support GQA requires only 30 minutes of additional autonomous agent effort, yielding up to 7.0\% performance improvement over cuDNN and 9.3\% over FlashAttention-4.
\zedit{In addition, we further test AVO on two structurally different kernels: GatedDeltaNet, a linear-attention variant, and Triangle Attention, a key operator in \cready{AlphaFold}; in both cases AVO produces kernels that consistently outperform expert-tuned baselines (Appendix~\ref{app:more_tasks}), demonstrating the generalizability of AVO across diverse kernel optimization tasks.
\cready{Finally, to isolate the contribution of the agentic variation operator itself, we benchmark AVO against three prior LLM-driven evolutionary search (EVO) baselines under a matched wall-clock budget on the same hardware, where AVO achieves an approximately $31\%$ improvement over the strongest EVO baseline (Appendix~\ref{app:benchmark_evo}).}}

Our contributions are as follows:
\begin{itemize}
    \item We introduce Agentic Variation Operators (AVO), a new family of evolutionary variation operators that elevate the agent from candidate generator to variation operator, autonomously exploring domain knowledge, implementing edits, and validating results through iterative interaction with the environment.
    \item We achieve state-of-the-art MHA throughput on NVIDIA B200 GPUs across the benchmarked configurations, reaching up to 1668 TFLOPS and outperforming cuDNN by up to 3.5\% and FlashAttention-4 by up to 10.5\%. Furthermore, we show that the discovered optimizations readily transfer to GQA, requiring only 30 minutes of autonomous adaptation and yielding gains of up to 7.0\% over cuDNN and 9.3\% over FlashAttention-4.
    \item We provide a detailed analysis of the micro-architectural optimizations discovered by the agent under the benchmarked settings, showing the agent performs genuine hardware-level reasoning rather than superficial code transformations.
    \zedit{\item We further demonstrate that AVO generalizes beyond attention to other widely-used GPU kernels, achieving up to $4.1\times$ speedup over Flash Linear Attention on GatedDeltaNet and consistent gains over NVIDIA's cuEquivariance on Triangle Attention.}
\end{itemize}

\section{Background}
\label{sec:background}

\subsection{Evolutionary Search and Variation Operators}
\label{sec:background_evo}

Evolutionary search optimizes over a space of candidates by maintaining a population $\mathcal{P}$ and iteratively expanding it with new solutions~\citep{back1997evolutionary}.
A population is a set of solution-score pairs $\mathcal{P} = \{(x_i, \mathbf{f}(x_i))\}$, where $\mathbf{f}$ is a scoring function that evaluates each candidate solution.
Each iteration produces a new candidate $x_{t+1}$ and updates the population:
\begin{equation}
\label{eq:evo_loop}
\mathcal{P}_{t+1} = \texttt{Update}\!\big(\mathcal{P}_t,\; (x_{t+1},\, \mathbf{f}(x_{t+1}))\big), \quad x_{t+1} = \texttt{Vary}(\mathcal{P}_t),
\end{equation}
where $\texttt{Update}$ adds the new solution to the population, possibly pruning low-score members to maintain a bounded archive.
We call $\texttt{Vary}$ the \emph{variation operator}: the mechanism by which new candidates are produced from existing ones.
In works such as FunSearch~\citep{funsearch2024}, AlphaEvolve~\citep{alphaevolve2025}, and related LLM-augmented evolutionary methods~\citep{lehman2022elm, ye2024reevo, chen2023evoprompting}, the variation operator decomposes into two stages:
\begin{equation}
\label{eq:classical_vary}
\texttt{Vary}(\mathcal{P}_t) = \texttt{Generate}\!\big(\texttt{Sample}(\mathcal{P}_t)\big),
\end{equation}
where $\texttt{Sample}$ selects one or more parent solutions from $\mathcal{P}_t$ (typically guided by score-based and diversity-based heuristics), and $\texttt{Generate}$ produces a new candidate conditioned on the sampled parents.

\paragraph{LLM-augmented variation.}
In these approaches, $\texttt{Generate}$ is implemented by an LLM that is prompted with the sampled parents and asked to produce a more optimized solution.
The $\texttt{Sample}$ step, however, remains a fixed algorithmic procedure: AlphaEvolve maintains an island-based evolutionary database inspired by MAP-Elites~\citep{mouret2015mapelites}, where a prompt sampler selects parent and inspiration programs using predefined fitness-based and diversity-based heuristics.
LoongFlow~\citep{wan2025loongflow} similarly relies on a MAP-Elites archive with Boltzmann selection for $\texttt{Sample}$, while structuring $\texttt{Generate}$ as a fixed Plan-Execute-Summarize pipeline where the LLM sequentially generates a modification plan, produces the code, and summarizes insights.
In all these approaches, the LLM only participates in $\texttt{Generate}$: the sampling strategy, evaluation protocol, population management, and the order of operations are all determined by the framework, not by the LLM.

\paragraph{Learned variation.}
TTT-Discover~\citep{yuksekgonul2025tttdiscover} goes further by updating the LLM policy itself through test-time gradient updates, enabling the model to learn an improved $\texttt{Generate}$ during the search.
Nevertheless, $\texttt{Sample}$ remains a fixed algorithm: a PUCT-based selection rule~\citep{silver2016alphago} determines which states to expand, and a buffer manages the population with predetermined update rules.
Even with a learned $\texttt{Generate}$, the LLM's role is still confined to candidate generation within a rigid algorithmic structure that prescribes when and how it is invoked.

In contrast, the agentic variation operator we introduce in Section~\ref{sec:method} replaces the entire $\texttt{Vary}$ with a self-directed agent that subsumes $\texttt{Sample}$, $\texttt{Generate}$, and evaluation into a single autonomous loop.
The agent has full agency over when to consult reference materials and past solutions $\mathcal{P}_t$, what diagnostic tests to run, and how to revise its optimization strategy.

AVO is orthogonal to the choice of population structure: the agentic operator can in principle be used within archive-based, island-based, or single-lineage evolutionary regimes.
In this paper we study the single-lineage setting to isolate the effect of the operator itself.

\subsection{Attention Kernels on Modern GPUs}
\label{sec:background_attention}

\paragraph{Attention computation.}
Given query, key, and value matrices $Q$, $K$, $V$, attention computes $O = \mathrm{softmax}(QK^\top / \sqrt{d})\, V$, where $d$ is the head dimension.
A naive implementation materializes the full $N \times N$ score matrix $S = QK^\top$, making the operation memory-bound for large sequence lengths $N$.
The FlashAttention algorithm~\citep{dao2022flashattention} avoids this by computing attention in tiles: it processes key blocks sequentially, maintaining a running softmax (with running row-maximum and row-sum) and accumulating the output $O$ incrementally.
This tiling eliminates the need to store the full score matrix, shifting the bottleneck from memory bandwidth to compute throughput on modern GPUs.

\paragraph{Attention kernel on Blackwell hardware.}
On NVIDIA's Blackwell architecture, state-of-the-art attention kernels such as FA4~\citep{zadouri2026flashattention4} employ \emph{warp specialization}: different warp groups within a thread block are assigned distinct roles in the attention pipeline.
\emph{MMA warps} execute the two core matrix multiplications via Blackwell's tensor core instructions: the QK GEMM (producing scores $S$) and the PV GEMM (multiplying the softmax output $P = \mathrm{softmax}(S)$ by $V$ to accumulate the output $O$).
\emph{Softmax warps} compute attention weights $P$ from the scores $S$, applying the online softmax algorithm with a running row-maximum.
\emph{Correction warps} rescale the output accumulator $O$ when the running maximum changes across K-block iterations (a requirement of the online softmax algorithm).
\emph{Load and epilogue warps} handle data movement via the Tensor Memory Accelerator (TMA).
In FA4's pipeline, these groups operate concurrently across two Q-tiles (a \emph{dual Q-stage} design), with barrier-based signaling to coordinate handoffs.
For causal attention, some K-block iterations are fully masked (no valid attention entries) and others are fully unmasked, leading to different execution paths within the same kernel.
With FA4 already representing a highly optimized design, further improvements demand deep hardware expertise, broad exploration across diverse optimization strategies, and repetitive debugging and profiling.

\section{Agentic Variation Operators}
\label{sec:method}

\begin{figure}[t]
  \centering
    \includegraphics[width=\textwidth]{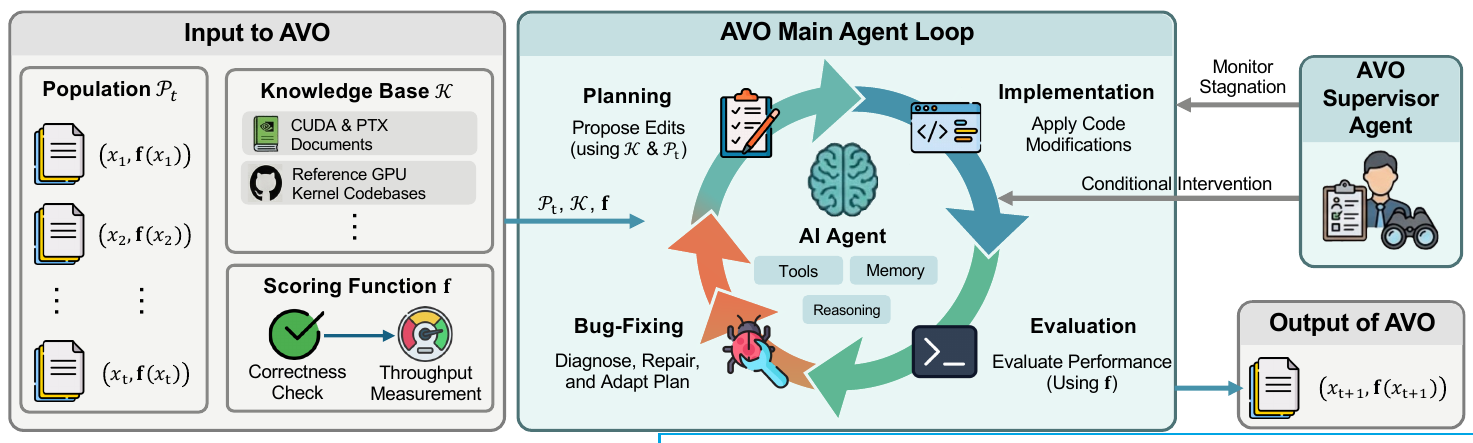}
  \caption{Illustration of the Agentic Variation Operator (AVO).}
  \label{fig:avo_overview}
\end{figure}

AVO consolidates the sampling, generation, and evaluation stages of evolutionary search into a single autonomous agent run, eliminating the rigid pipeline that constrains existing approaches.
Below we formalize this operator, detail what occurs within a single variation step, and describe the mechanism that enables multi-day autonomous exploration.

\subsection{Formulation}
\label{sec:method_formal}

Previous evolutionary search approaches~\citep{funsearch2024, alphaevolve2025} decompose the variation operator as:
\begin{equation}
\label{eq:evo}
\texttt{Vary}(\mathcal{P}_t) = \texttt{Generate}(\texttt{Sample}(\mathcal{P}_t)),
\end{equation}
confining the LLM to the $\texttt{Generate}$ step within a fixed pipeline.
As illustrated in Figure~\ref{fig:avo_overview}, AVO replaces this decomposition with a single autonomous agent run:
\begin{equation}
\label{eq:avo}
\texttt{Vary}(\mathcal{P}_t) = \texttt{Agent}(\mathcal{P}_t, \mathcal{K}, \mathbf{f}),
\end{equation}
where $\mathcal{P}_t = \{(x_1, \mathbf{f}(x_1)), \ldots, (x_t, \mathbf{f}(x_t))\}$ is the full lineage of solutions and their scores, $\mathcal{K}$ is a domain-specific knowledge base, and $\mathbf{f}$ is the scoring function.

In our setting, each $x_i$ is a CUDA kernel implementation (source code with inline PTX), and $\mathbf{f}$ evaluates a candidate along two dimensions: numerical correctness against a reference implementation, and throughput in TFLOPS on the target hardware. In practice, $\mathbf{f}(x_i) = (f_1(x_i), f_2(x_i), \ldots, f_n(x_i))$ is an $n$-dimensional vector and $f_j$ represents the score for test configuration $j$. 
A candidate $x_i$ that fails correctness is assigned zero score (i.e., $f_j(x_i) = 0$) regardless of throughput.
The knowledge base $\mathcal{K}$ contains CUDA programming guides, PTX ISA documentation, Blackwell architecture specifications, and existing kernel implementations including FlashAttention-4 source code.

AVO defines a family of agentic variation operators for evolutionary search. In this work, we instantiate AVO in a single-lineage autonomous run starting from an initial program $x_0$, producing a sequence of committed improvements $x_1, x_2, \ldots, x_t$. The accumulated lineage $\mathcal{P}_t$ serves as context for subsequent variation steps.

\subsection{Anatomy of a Variation Step}
\label{sec:method_step}

A single variation step in AVO, producing $x_{t+1}$ from the current lineage $\mathcal{P}_t$, is an autonomous agent loop.
The agent is a general-purpose coding agent with planning, tool use, and persistent memory (details in Section~\ref{sec:experiments}), and a single step may involve numerous internal actions.

We observe that the agent frequently examines multiple prior implementations in $\mathcal{P}_t$ within a single variation step, comparing their profiling characteristics to identify bottlenecks and opportunities, and consulting documentation in $\mathcal{K}$ to understand the relevant hardware constraints before implementing a candidate optimization.
The agent then invokes $\mathbf{f}$ to test the result.
When a candidate fails correctness checks or fails to improve on the current benchmark suite, the agent diagnoses the issue and revises its approach, repeating this edit-evaluate-diagnose cycle until it commits a satisfactory $x_{t+1}$.
This design allows the agent to adapt its optimization strategy as the search progresses: early steps may focus on structural changes informed by reference implementations in $\mathcal{K}$, while later steps can shift toward micro-architectural tuning guided by profiling feedback from $\mathbf{f}$ and patterns observed across the accumulated lineage $\mathcal{P}_t$.

In our current implementation, we persist a new committed version only when it passes correctness checks and matches or improves the benchmark score relative to the best committed version so far; unsuccessful intermediate attempts remain part of the agent’s internal search trajectory but are not added to the committed lineage.

\subsection{Continuous Evolution}
\label{sec:method_loop}

Although AVO is defined at the level of variation operators for evolutionary search, the present study evaluates a single-lineage continuous instantiation, leaving a full treatment of population-level branching and archive management to future work; \crebut{we report a preliminary cross-lineage study in Appendix~\ref{app:population}}.
The AVO agent operates as a continuous loop that periodically produces new solutions without human intervention.
Each committed version $x_i$ is persisted as a git commit along with its score, maintaining full state continuity across the entire evolutionary process.

In long-running autonomous optimization, two failure modes can impede progress: the agent may \emph{stall} when it exhausts its current line of exploration, or it may enter \emph{unproductive cycles} of edits that repeatedly fail to improve scores.
To mitigate both, AVO incorporates a self-supervision mechanism that detects these scenarios and intervenes. Once triggered, the mechanism reviews the overall evolutionary trajectory and steers the search toward several candidate optimization directions. This conditional intervention effectively redirects exploration with fresh perspective when the current strategy has plateaued.

\crebut{Concretely, each run maintains the archive $\mathcal{P}_t$ of committed implementations and their measured fitness on the local filesystem, which the agent may consult freely when producing a new candidate.}
The 7-day run that produced our final multi-head attention kernel spanned 40 successive versions. Throughout this process, the main agent autonomously decided when to attempt new optimizations, when to revisit earlier approaches in $\mathcal{P}_t$, and when to shift strategy, while the supervisor maintained forward progress by intervening during periods of stagnation.

\section{Experiments}
\label{sec:experiments}

\zedit{This section reports our results on multi-head attention and grouped-query attention kernels. Additional applications of AVO to other kernel optimization tasks and comparisons against existing evolutionary search (EVO) \cready{baselines} are provided in Appendix~\ref{app:more_tasks} and Appendix~\ref{app:benchmark_evo}, respectively.}

\subsection{Setup}
\label{sec:setup}

\paragraph{Agent.}
We use an internally-developed general-purpose coding agent powered by \zedit{Claude Opus 4.6} as the AVO variation operator.
The agent has access to standard software engineering tools, including autonomous code editing, shell command execution, file system navigation, and documentation retrieval.
It maintains persistent memory through its conversation history, which accumulates the full context of prior edits, compiler outputs, profiling results, and reasoning across the evolutionary process.
No task-specific modifications are made to the agent for kernel optimization; the same agent used for general software engineering tasks is deployed here, with the domain-specific knowledge base $\mathcal{K}$ and scoring function $\mathbf{f}$ provided to the agent as described in Section~\ref{sec:method_formal}.

\paragraph{Hardware and software.}
Following the setup of FA4~\citep{zadouri2026flashattention4}, all \cready{experiments in this section} are conducted on NVIDIA B200 GPUs with CUDA 13.1 and PyTorch 2.10.0.

\paragraph{Baselines.}
We compare against two state-of-the-art baselines:
(1) \textbf{cuDNN}: NVIDIA's closed-source attention kernel, measured using cuDNN version 9.19.1, which includes custom optimizations for Blackwell; and
(2) \textbf{FlashAttention-4} (FA4)~\citep{zadouri2026flashattention4}: the latest open-source attention kernel optimized for Blackwell, measured using the official implementation (commit \texttt{71bf77c}).

\paragraph{Benchmark Configurations.}
We evaluate the forward prefilling throughput with head dimension 128 and BF16 precision across sequence lengths $\{4096, 8192, 16384, 32768\}$.
Following FlashAttention-4~\citep{zadouri2026flashattention4}, we control the total number of tokens to 32768 by adjusting the batch size for each sequence length (e.g., batch size 8 at sequence length 4096, batch size 1 at sequence length 32768).
For multi-head attention (MHA), we use 16 heads under both causal and non-causal masking.
For grouped-query attention (GQA), we evaluate two configurations drawn from the Qwen3 model family~\citep{yang2025qwen3}: 32 query heads with 4 KV heads (group size 8, as in Qwen3-30B-A3B) and 32 query heads with 8 KV heads (group size 4, as in Qwen3-8B).
For throughput measurement, we \cready{use} the same timing script from the FA4 repository\footnote{\url{https://github.com/Dao-AILab/flash-attention/blob/main/benchmarks/benchmark_attn.py}, BSD-3-Clause license.} and the same number of warm-up and repeat rounds as the FA4 paper. In addition, we \cready{run each experiment} 10 times to obtain the average performance and the standard deviation.
The same setup is used both for agent evolution and for benchmarking the final evolved kernels against the baselines.

\begin{figure}[t]
  \centering
  \includegraphics[width=\textwidth]{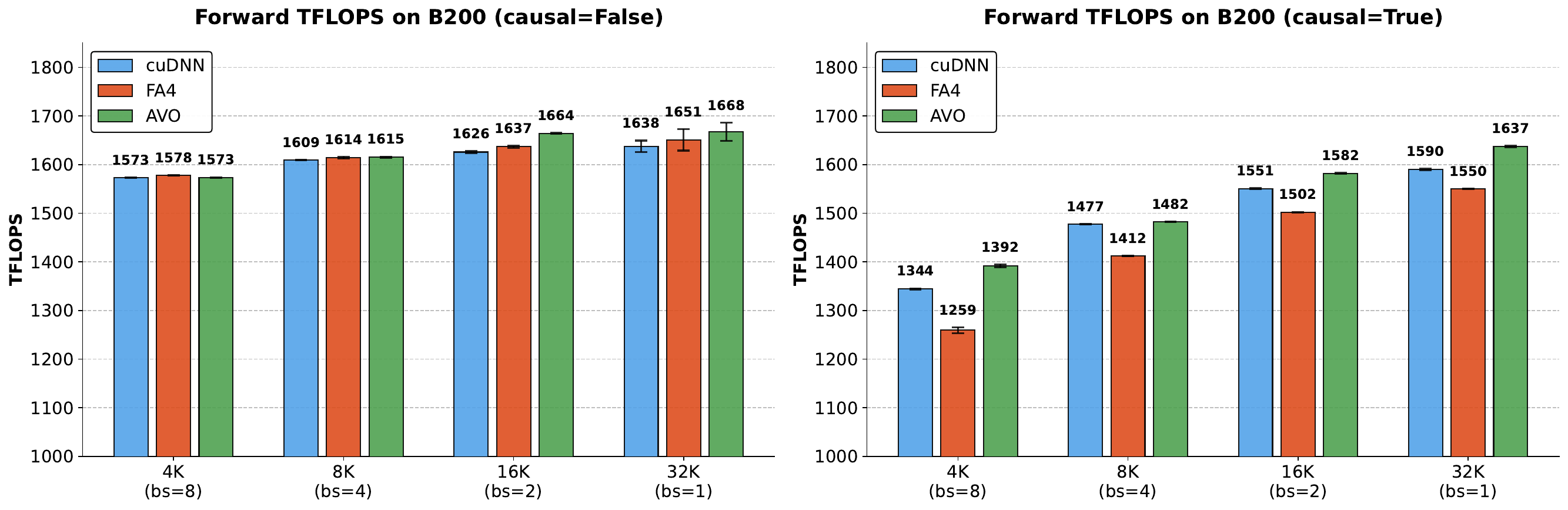}
  \caption{Multi-head attention forward-pass prefilling throughput (TFLOPS) on NVIDIA B200 with head dimension 128, 16 heads, and BF16 precision. Batch size and sequence length are varied with a fixed total of 32k tokens.}
  \label{fig:mha_results}
\end{figure}

\subsection{Multi-Head Attention}
\label{sec:results_mha}

Figure~\ref{fig:mha_results} presents the benchmarking results for MHA.
On causal attention, AVO outperforms both baselines across all tested configurations, with gains ranging from $+0.4\%$ to $+3.5\%$ over cuDNN and $+5.0\%$ to $+10.5\%$ over FA4.
On non-causal attention, AVO achieves modest gains at longer sequences ($+1.8\%$ to $+2.4\%$ over cuDNN at sequence lengths larger than 16384) but is within measurement noise of both baselines at shorter sequences. In Section~\ref{sec:results_trajectory}, we show how the agent obtains the performance gains through continuous evolution.

\subsection{Grouped-Query Attention}
\label{sec:results_gqa}

To evaluate whether agent-discovered optimizations transfer beyond the benchmark settings used in evolution, we prompted the AVO agent to adapt the evolved MHA kernel to support GQA.
The agent completed this adaptation autonomously in approximately 30 minutes, producing a GQA-capable kernel without any human guidance on the required changes.

\begin{figure}[t]
  \centering
  
   \includegraphics[width=\textwidth]{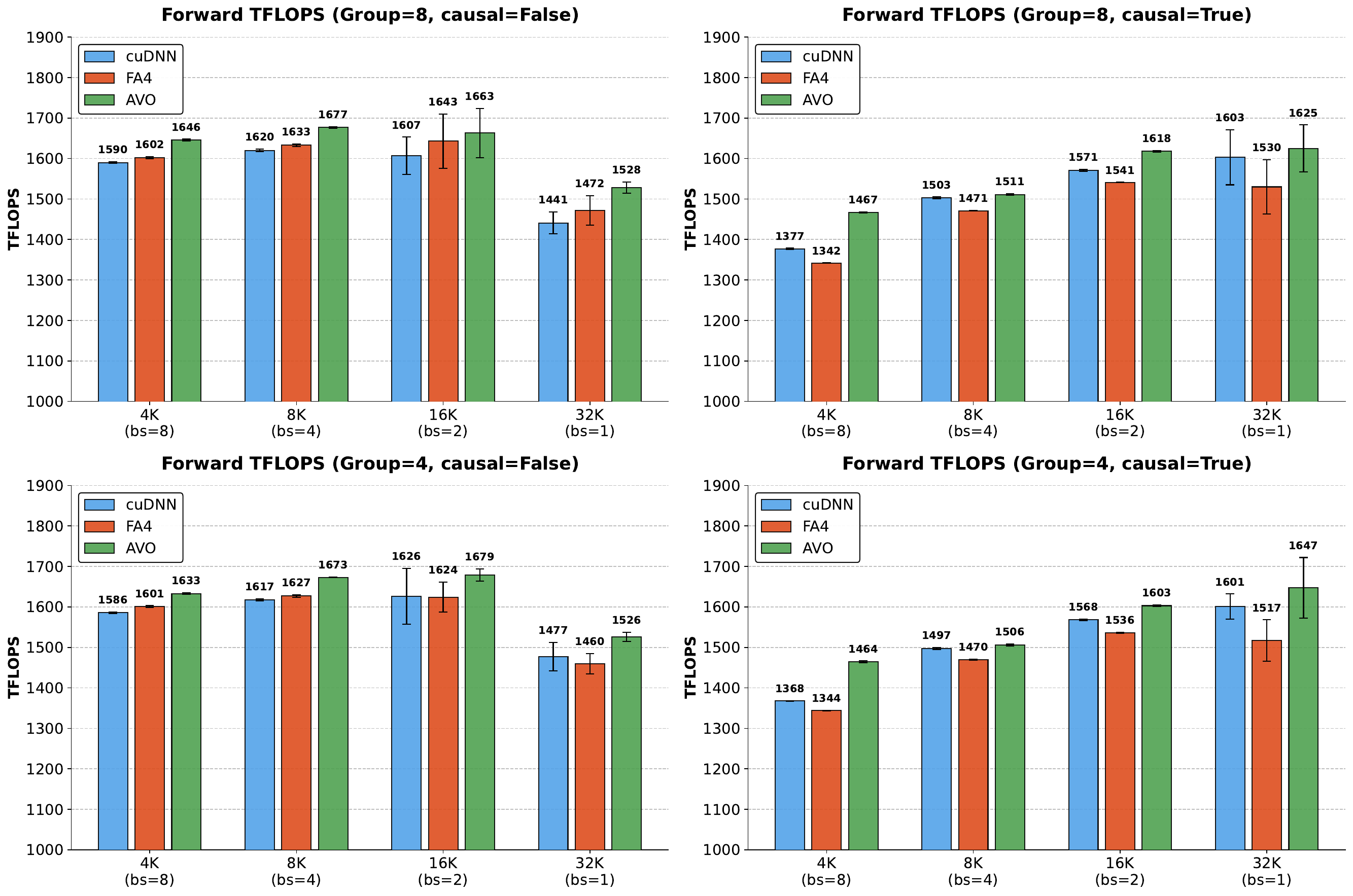}
  \caption{Grouped-query attention forward-pass prefilling throughput (TFLOPS) on NVIDIA B200 with 32 query heads, head dimension 128 and BF16 precision. Results are shown for two GQA configurations (group sizes 8 and 4) under both causal and non-causal masking. The GQA kernel was produced by prompting the AVO agent to adapt the evolved MHA kernel, requiring approximately 30 minutes of autonomous effort.}
  \label{fig:gqa_results}
\end{figure}

Figure~\ref{fig:gqa_results} presents the results across two GQA configurations.
AVO outperforms both baselines across all configurations.
On causal GQA, AVO achieves up to $+7.0\%$ over cuDNN and $+9.3\%$ over FA4.
On non-causal GQA, gains reach up to $+6.0\%$ over cuDNN and $+4.5\%$ over FA4.
The strong GQA performance demonstrates that the optimizations discovered by the agent during MHA evolution are not specific to the MHA configurations used during evolution, but generalize to the distinct compute and memory access patterns of GQA.

\subsection{Evolution Trajectory}
\label{sec:results_trajectory}

\begin{figure}[t]
  \centering
  \includegraphics[width=\textwidth]{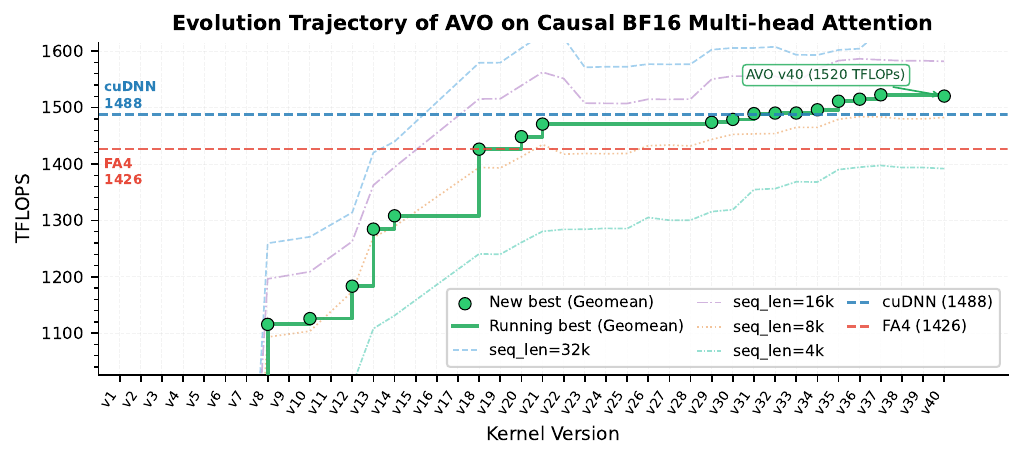}
  \vspace{-2em}
  \caption{Evolution trajectory of AVO across 40 kernel versions over 7 days on \textbf{causal} MHA. The solid green line tracks the running-best geometric mean throughput across all configurations; green circles mark versions that set a new best. Dashed colored lines show per-configuration throughput (seq\_len = 4k, 8k, 16k, 32k). Horizontal dashed lines indicate the geometric mean throughput of cuDNN and FA4.}
  \label{fig:evolution_causal_trajectory}
\end{figure}

\begin{figure}[t]
  \centering
  \includegraphics[width=\textwidth]{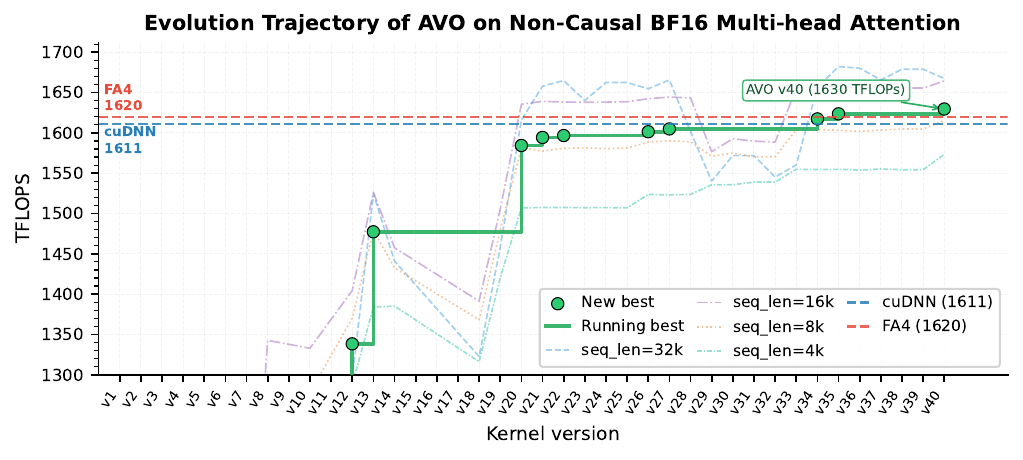}
  \vspace{-2em}
  \caption{Evolution trajectory of AVO across 40 kernel versions over 7 days on \textbf{non-causal} MHA. The solid green line tracks the running-best geometric mean throughput across all configurations; green circles mark versions that set a new best. Dashed colored lines show per-configuration throughput (seq\_len = 4k, 8k, 16k, 32k). Horizontal dashed lines indicate the geometric mean throughput of cuDNN and FA4.}
  \label{fig:evolution_noncausal_trajectory}

\end{figure}

In Figure~\ref{fig:evolution_causal_trajectory} and Figure~\ref{fig:evolution_noncausal_trajectory}, we show the evolution trajectory of AVO across the 40 committed kernel versions produced during the 7-day evolution.
Note that these trajectories visualize the committed sequence, rather than the full internal search tree explored between the commits.
We \cready{observe} the following patterns:

\paragraph{Scale of exploration.}
The 40 committed versions shown in the trajectory represent only the successful outcomes of a much larger search.
Over the 7-day evolution, the agent explored over 500 candidate optimization directions internally, including attempts that failed correctness checks, regressed throughput, or were abandoned after profiling.
This volume of systematic exploration, each direction requiring reading documentation, implementing changes, compiling, testing, and profiling, far exceeds what a human engineer could accomplish in the same timeframe.

\paragraph{Discrete jumps rather than gradual improvement.}
Throughput improves in distinct steps separated by plateaus where successive versions refine implementation details without measurably changing performance.
The five largest gains correspond to architectural inflection points: the introduction of QK-PV interleaving with bitmask causal masking (version 8), a restructured single-pass softmax computation (version 13), the branchless accumulator rescaling with a lighter memory fence for unmasked iterations (version 20), the correction/MMA pipeline overlap (version 30), and register rebalancing across warp groups (version 33). We discuss some of the representative optimizations in Section~\ref{sec:analysis}.
The remaining versions contribute individually smaller but collectively substantial micro-architectural refinements.

\section{Analysis of Agent-Discovered Optimizations}
\label{sec:analysis}

The 40-version AVO evolution produced multi-level optimizations that individually yield measurable throughput gains and collectively account for the improvements reported in Section~\ref{sec:experiments}.
We examine \crebut{three} representative optimizations to illustrate the nature and depth of the agent's hardware reasoning.
For each, we describe the bottleneck the agent identified in its own kernel, the change it made, and its measured impact (ablation between the version immediately before and after).
\crebut{Table~\ref{tab:opt_summary} summarizes their measured impact.}

\begin{crebutsec}
\begin{table}[t]
  \caption{Summary of the agent-discovered optimizations examined in this section and their measured ablation gains (geomean throughput improvement over the immediately preceding version, across all benchmark configurations).}
  \label{tab:opt_summary}
  \centering
  \small
  \begin{tabular}{llcc}
    \toprule
    Optimization & Versions & Non-causal & Causal \\
    \midrule
    Branchless accumulator rescaling        & v19 $\to$ v20 & $+8.1\%$ & $+1.6\%$ \\
    Correction/MMA pipeline overlap         & v29 $\to$ v30 & $+1.1\%$ & $+0.4\%$ \\
    Register rebalancing across warp groups & v32 $\to$ v33 & $+2.1\%$ & $\sim 0\%$ \\
    \bottomrule
  \end{tabular}
\end{table}
\end{crebutsec}

\begin{crebutsec}
\subsection{Branchless Accumulator Rescaling}
\label{sec:analysis_branchless}

\paragraph{Bottleneck.}
The online softmax algorithm requires the output accumulator $O$ to be rescaled whenever the running row-maximum changes across key blocks (Section~\ref{sec:background_attention}).
In version 19 of the AVO kernel, this rescaling was guarded by a conditional branch that checked whether any thread in the warp required it and skipped the operation when the maximum was unchanged.
While this avoids unnecessary computation, the branch introduces warp synchronization overhead on every iteration of the key-block loop, and the conditional control flow prevents the use of lighter memory fences in the correction path.

\paragraph{AVO's approach.}
In version 20, the agent replaced the conditional branch with a branchless speculative path: the rescale factor is always computed, and a predicated select substitutes $1.0$ when rescaling is unnecessary, at a cost negligible compared to the synchronization overhead it replaces.
Eliminating the branch also removed warp divergence in the correction path, allowing a blocking memory fence to be replaced with a lighter non-blocking fence that merely enforces ordering; this is safe because uniform control flow guarantees reconvergence before the next synchronization point.
Together these changes yield $+8.1\%$ geomean throughput on non-causal and $+1.6\%$ on causal attention, the largest single optimization in the evolution.
The asymmetry arises because the branchless path applies only to fully unmasked iterations: non-causal attention processes all key blocks without masking, while causal attention retains the branched logic for masked blocks.
\end{crebutsec}

\subsection{Correction/MMA Pipeline Overlap}
\label{sec:analysis_split}

\paragraph{Bottleneck.}
The attention pipeline processes two Q-tiles concurrently (dual Q-stage; see Section~\ref{sec:background_attention}), each requiring a PV GEMM followed by output normalization by the correction warp.
In version 29 of the AVO kernel, the two stages were serialized at the MMA-to-correction boundary: the correction warp had to wait for both PV GEMMs to complete before it could begin normalizing either stage's output, leaving it idle throughout the second GEMM.

\paragraph{AVO's approach.}
In version 30, the agent restructured the pipeline to allow the correction warp to begin normalizing the first stage's output as soon as its PV GEMM completes, overlapping this work with the second stage's PV GEMM.
This converts a sequential dependency into a pipelined execution, reducing the idle time on the correction warp and yielding $+1.1\%$ geomean throughput on non-causal and $+0.4\%$ on causal attention.

\subsection{Register Rebalancing Across Warp Groups}
\label{sec:analysis_registers}

\paragraph{Bottleneck.}
Blackwell partitions a fixed budget of 2048 warp-registers per SM across warp groups.
In version 32 of the AVO kernel, the allocation followed the pattern of FlashAttention-4~\citep{zadouri2026flashattention4}: 192 registers for the 8 softmax warps, 80 for the 4 correction warps, and 48 for the remaining 4 warps.
Profiling revealed that the correction warp group was spilling values to slower local memory due to its limited 80-register budget, while the softmax group had substantial headroom.

\paragraph{AVO's approach.}
In version 33, the agent redistributed 8 registers from the softmax group to each of the other two groups, arriving at a 184/88/56 allocation.
This redistribution is viable because the AVO kernel's softmax implementation processes score values in small fragments with packed arithmetic, resulting in a low peak register usage that leaves ample headroom even at 184 registers.
The correction warp group benefits from the additional registers because, following the pipeline overlap optimization (Section~\ref{sec:analysis_split}), it runs concurrently with the second PV GEMM and is on the execution critical path.
With 88 rather than 80 registers, fewer output values spill to local memory, reducing stalls and yielding $+2.1\%$ geomean throughput on non-causal attention (within noise on causal).

\begin{crebutsec}
\subsection{Discussion}
\label{sec:analysis_discussion}
What is notable about these optimizations is that each requires jointly reasoning about multiple hardware subsystems, including synchronization and memory ordering, pipeline scheduling, and register allocation, rather than tuning any single parameter in isolation.
This depth of reasoning, carried out autonomously through iterative interaction with documentation and profiling feedback, suggests that agentic variation operators can serve as an effective mechanism for expert-level kernel optimization.
\end{crebutsec}

\section{Conclusion}
\label{sec:conclusion}

We introduced Agentic Variation Operators (AVO), a new family of evolutionary variation operators that elevate the agent from candidate generator to variation operator.
Applied to forward-pass attention on NVIDIA Blackwell GPUs, AVO produces kernels surpassing cuDNN by up to 3.5\% and FlashAttention-4 by up to 10.5\% over 7 days of continuous autonomous evolution.
Furthermore, we show that the discovered optimizations transfer readily to grouped-query attention, requiring only 30 minutes of additional autonomous adaptation.
\cready{Beyond attention, AVO proves effective across a broad range of GPU kernels, outperforming the expert-tuned reference on every additional kernel class we evaluate (Appendix~\ref{app:more_tasks}).}
Together, these results demonstrate that AVO can discover performance-critical micro-architectural optimizations that produce kernels surpassing state-of-the-art expert-engineered implementations.
Because AVO operates at the level of variation operators rather than being tied to a specific domain,  it points toward a broader path for autonomous optimization beyond attention kernels, including other performance-critical software systems on diverse hardware platforms, and engineering or scientific domains that demand extended autonomous exploration.

\section*{Acknowledgement}
We thank the NVIDIA Cutlass, cuDNN, TensorRT-LLM, FlashInfer, DevTech, IPP, and Compiler teams for valuable feedback and support.
We also thank the FlashAttention-4 authors for open-sourcing their implementation and benchmark scripts, which served as a baseline and a reference for this work.
\cready{This work was funded and supported by NVIDIA.}

\newpage

\bibliographystyle{unsrtnat}
\bibliography{references}

\appendix

\section{Generalization to Additional Kernel Optimization Tasks}
\label{app:more_tasks}

\crebut{Beyond the dense MHA and GQA prefill kernels of Section~\ref{sec:experiments}, we evaluate AVO on two further groups of kernels: linear and structured attention operators (Appendix~\ref{app:more_tasks_linear}), and sparse attention and non-attention kernels (Appendix~\ref{app:more_tasks_sparse}).}

\subsection{Linear and Structured Attention}
\label{app:more_tasks_linear}

\begin{figure}[h]
  \centering
  \includegraphics[width=\textwidth]{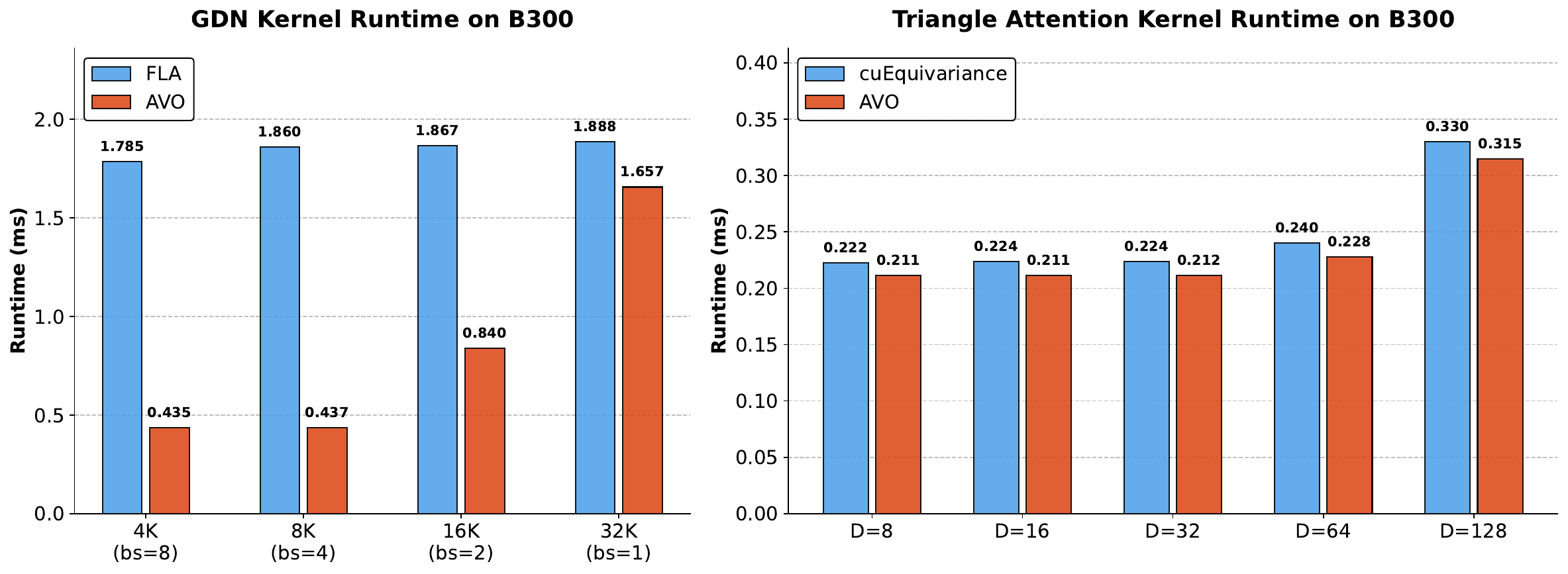}
  \caption{Kernel runtime (ms) on NVIDIA Blackwell B300 GPUs, comparing AVO against Flash Linear Attention (FLA) for GatedDeltaNet and against NVIDIA's cuEquivariance for Triangle Attention. \textbf{Left:} GatedDeltaNet (GDN) forward-pass runtime under varying sequence lengths and batch sizes (maintaining a constant 32K total tokens). \textbf{Right:} Triangle Attention runtime across different hidden dimensions ($D=8$ to $128$) with a fixed sequence length of 512 and 4 heads.}
  \label{fig:gdn_triattention}
\end{figure}

\paragraph{Tasks and Baselines.} We further apply AVO to two widely used GPU kernels: (1) GatedDeltaNet~\cite{yang2025gdn}, a recent linear attention operator serving as an efficient alternative to traditional softmax attention computation; and (2) Triangle Attention, introduced in AlphaFold2~\cite{jumper2021highly} to model the geometric relationships between interacting amino acid pairs. We use expert-optimized libraries as our performance baselines: Flash Linear Attention (FLA)~\cite{yang2024fla} for GatedDeltaNet, and NVIDIA's cuEquivariance~\cite{cuEquivariance} for Triangle Attention.

\paragraph{Agent and Hardware.} We use the same agent setup, time budget (i.e., 7 days per task), and benchmarking protocol as described in Section~\ref{sec:experiments}. All evaluations are conducted on NVIDIA Blackwell B300 GPUs (rather than the B200 used in the main experiments).

\paragraph{Benchmark Configurations.} For GatedDeltaNet, we benchmark using the configuration of Qwen3-Next-80B-A3B~\cite{cao2026qwen3codernexttechnicalreport} (matching the head count and head dimension), and test performance under the same sequence length and batch size combinations as in Section~\ref{sec:experiments}. For Triangle Attention, we fix the sequence length at 512 and the head count at 4 to mirror realistic biological workloads and the AlphaFold2~\cite{jumper2021highly} architecture, respectively. We sweep the head dimension ($D$) from 8 to 128 to systematically evaluate the kernel's performance scaling from memory-bound to compute-bound regimes.  

\paragraph{Results.} As shown in Figure~\ref{fig:gdn_triattention}, AVO generalizes to both kernels and yields substantial gains over the expert baselines. For GatedDeltaNet (Figure~\ref{fig:gdn_triattention}, left), the AVO-generated kernel substantially reduces runtime compared to FLA across all sequence lengths. Most notably, at shorter sequence lengths with larger batch sizes (e.g., 4K sequence length, batch size 8), AVO achieves a $4.1\times$ speedup, reducing latency from 1.785~ms to 0.435~ms. For Triangle Attention (Figure~\ref{fig:gdn_triattention}, right), AVO-generated kernels consistently outperform NVIDIA's cuEquivariance library across all tested hidden dimensions, with a stable runtime reduction of roughly 4.7\% to 5.5\% across the $D=8$ to $128$ sweep. These results demonstrate AVO's capability to discover performance optimizations even when competing against vendor-engineered libraries.

\begin{crebutsec}
\subsection{Sparse Attention and Non-Attention Kernels}
\label{app:more_tasks_sparse}

\paragraph{Tasks and Baselines.} We apply AVO to three further kernel classes: (1) sparse multiple-sequence-alignment (MSA) attention, benchmarked against the official MiniMax implementation; (2) fused FP8 mixture-of-experts (MoE), benchmarked against the FlashInfer-Bench~\citep{xing2026flash} reference; and (3) MXFP4 grouped GEMM, benchmarked against the OpenAI Triton reference. Sparse MSA extends the evaluation from dense to sparse attention, while fused MoE and grouped GEMM fall outside attention entirely.

\paragraph{Agent and Hardware.} Each task is evolved for 24 hours on a single NVIDIA B200, using the same agent setup and benchmarking protocol as Section~\ref{sec:experiments}. No task-specific changes are made to the search procedure; adapting AVO to a new task requires only supplying the corresponding correctness and performance benchmark.

\paragraph{Results.} As shown in Table~\ref{tab:new_classes}, AVO outperforms the expert-engineered reference on all three kernel classes. Together with the linear and structured attention results above, this indicates that the agentic variation operator transfers across kernel classes rather than being specialized to attention.

\begin{table}[h]
  \caption{AVO on sparse attention and non-attention kernels. Speedup is the geometric mean over the listed workloads relative to the expert-engineered reference.}
  \label{tab:new_classes}
  \centering
  \footnotesize
  \begin{tabular}{lllc}
    \toprule
    Kernel class & Reference & Workloads & Speedup \\
    \midrule
    Fused FP8 MoE      & FlashInfer-Bench~\citep{xing2026flash} & 19 traces; len.\ 1--14{,}107        & $1.604\times$ \\
    MXFP4 grouped GEMM & OpenAI Triton                          & 5 layers $\times$ 4 token counts     & $1.102\times$ \\
    Sparse MSA         & MiniMax MSA                            & batch 1--4; len.\ 4{,}096--32{,}768  & $1.492\times$ \\
    \bottomrule
  \end{tabular}
\end{table}
\end{crebutsec}

\section{Comparison with EVO Baselines}
\label{app:benchmark_evo}

To compare AVO against prior LLM-driven evolutionary search frameworks, we evaluate on three kernel optimization tasks adopted from K-Search~\cite{cao2026k}, drawn from the FlashInfer-Bench~\cite{xing2026flash} dataset: \textbf{GQA decode}, \textbf{MLA decode}, and \textbf{MLA prefill}. All experiments are conducted on a server with a single NVIDIA H100 GPU.

\paragraph{Baselines.} We compare AVO against three representative LLM-based evolutionary search (EVO) baselines: K-Search~\cite{cao2026k}, OpenEvolve~\cite{openevolve} (an open-source implementation of AlphaEvolve~\cite{alphaevolve2025}), and ShinkaEvolve~\cite{lange2025shinka}. We follow the same benchmarking script and evaluation protocol as K-Search; the speedup of each generated kernel is measured against the FlashInfer-Bench~\cite{xing2026flash} reference implementation.

\paragraph{AVO setup.} We use the same agentic setup as described in Section~\ref{sec:setup}. For a fair comparison with the EVO baselines, we cap the AVO evolution time budget per task at the wall-clock time required by K-Search, the most time-consuming baseline, on the same server, which averages $11.22$ hours per task across the three benchmarks.

\begin{figure}[t]
  \centering
  \includegraphics[width=\textwidth]{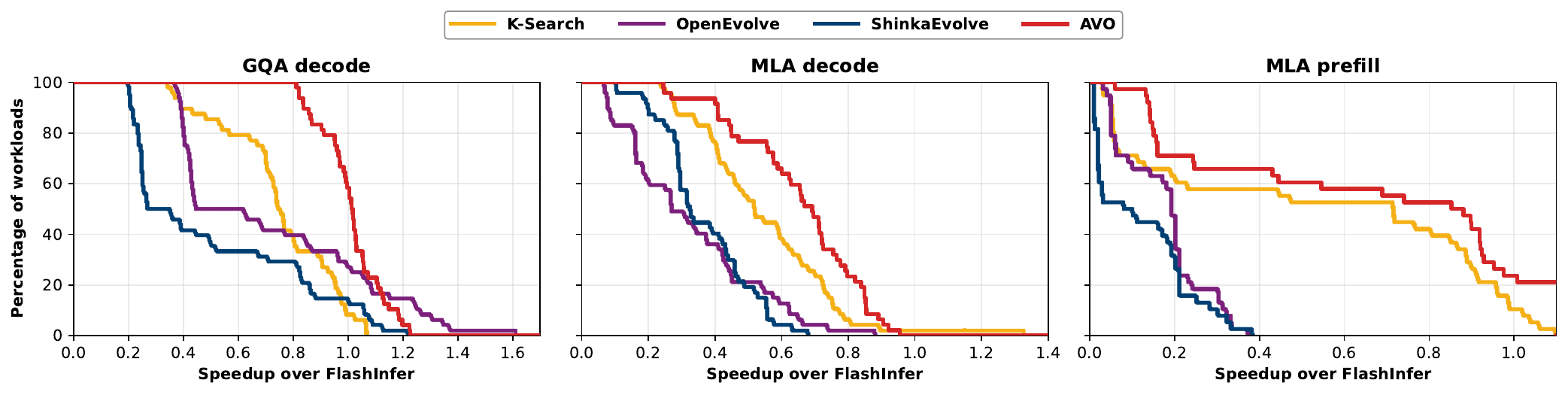}
  \caption{Time-matched comparison between AVO and prior LLM-driven evolutionary search frameworks (K-Search, OpenEvolve, ShinkaEvolve) on three FlashInfer-Bench tasks. Each curve shows the percentage of workloads (y-axis) achieving at least the corresponding speedup over the FlashInfer-Bench reference implementation (x-axis); curves further to the upper right indicate stronger and more uniform improvements. All methods are run for $11.22$ hours per task on a single NVIDIA H100 GPU.}
  \label{fig:avo_evo_survival}
\end{figure}

\paragraph{Results.} Figure~\ref{fig:avo_evo_survival} compares AVO-generated kernels against those produced by the EVO baselines. On all three tasks, AVO outperforms the strongest EVO baseline (K-Search) by $33.0\%$, $38.6\%$, and $22.0\%$ on GQA decode, MLA decode, and MLA prefill respectively, corresponding to a geometric-mean improvement of \cready{approximately} $31\%$. Beyond the comparison with EVO baselines, AVO further identifies optimization opportunities that surpass the FlashInfer-Bench reference: for example, AVO is faster than the reference on $60.4\%$ of GQA decode workloads and $10.5\%$ of MLA prefill workloads. Together, these results confirm AVO's consistent advantage over prior EVO baselines \cready{under a matched wall-clock budget on the same hardware}, isolating the variation operator itself as the source of the gains, and further show that AVO can discover non-trivial optimizations that match or exceed expert-engineered references on a meaningful fraction of workloads.

\section{Comparison Using FA4-Reported Baseline Performance}
\label{app:fa4_baselines}

\begin{figure}[h]
  \centering
  \includegraphics[width=\textwidth]{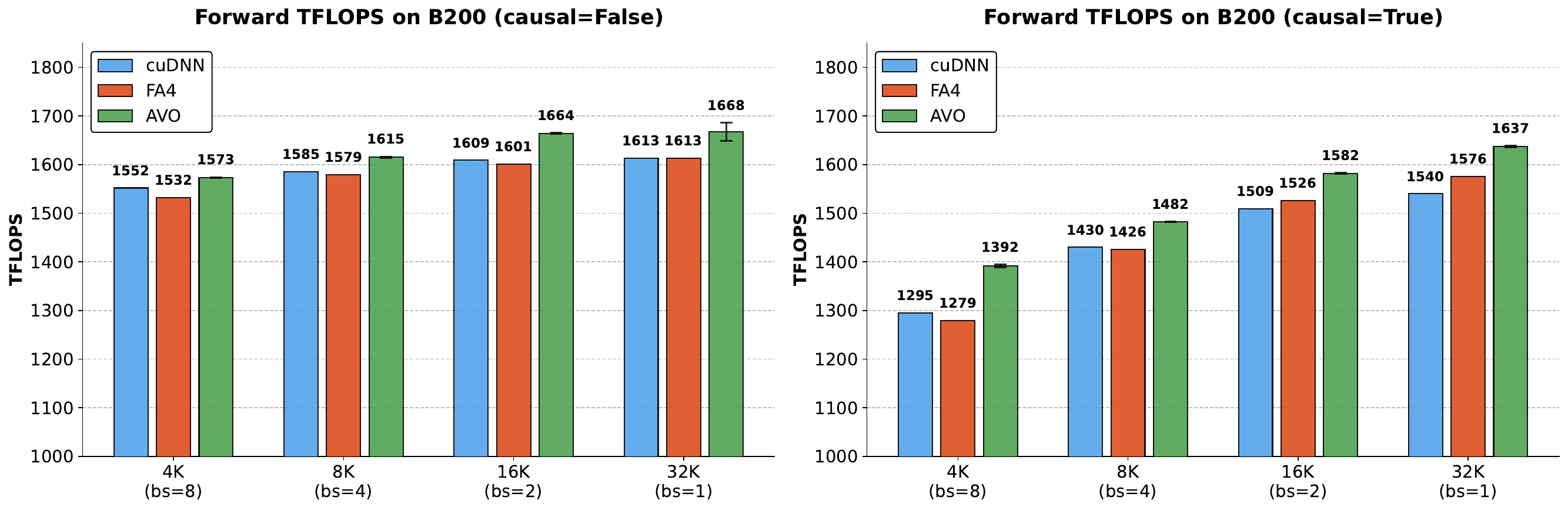}
  \caption{Multi-head attention forward-pass throughput (TFLOPS) on NVIDIA B200, comparing AVO (measured on our hardware) against cuDNN and FA4 baseline numbers as reported in the FA4 paper~\citep{zadouri2026flashattention4}. Head dimension 128, 16 heads, BF16. Left: non-causal. Right: causal.}
  \label{fig:mha_official}
\end{figure}

Section~\ref{sec:experiments} reports cuDNN and FA4 throughput measured on our hardware. In practice, minor system-level differences (driver versions, thermal conditions, clock frequencies) can affect absolute TFLOPS. Therefore, we additionally compare AVO against the cuDNN and FA4 numbers published in the FA4 paper~\citep{zadouri2026flashattention4}. Figure~\ref{fig:mha_official} presents this comparison.

On non-causal attention, AVO outperforms the FA4-reported baselines across all configurations, with gains of $+1.4\%$ to $+3.4\%$ over cuDNN and $+2.3\%$ to $+3.9\%$ over FA4.
On causal attention, AVO achieves $+3.6\%$ to $+7.5\%$ over cuDNN and $+3.7\%$ to $+8.8\%$ over FA4, with the largest gains observed at shorter sequences (bs=8, seq=4096).
These results are broadly consistent with the comparisons in Section~\ref{sec:experiments}.

\section{Generalization to Unseen Workload Shapes}
\label{app:shapes}

\begin{crebutsec}
Appendix~\ref{app:more_tasks} asks whether AVO transfers to new kernels; here we ask whether the evolved kernels transfer to workload shapes they were never tuned on. The benchmark suite of Section~\ref{sec:experiments}---four sequence lengths under both causal and non-causal masking---also defined the scoring function during evolution, so we expand the sweep to batch sizes $\{1, 2, 4, 8, 16\}$, sequence lengths $\{1024, 2048, 4096, 8192, 16384, 32768, 65536, 131072\}$, and both causal and non-causal masking, yielding 80 configurations per kernel family. Both kernels are evaluated on an NVIDIA B200 using the same correctness and timing protocol as Section~\ref{sec:setup}. Improvements are reported as the geometric mean of baseline latency divided by AVO latency, and ``wins'' counts the configurations on which AVO attains lower latency.

\begin{table}[h]
  \caption{AVO across 80 configurations per kernel family (batch size $\times$ sequence length $\times$ mask), well beyond the benchmark suite used during evolution.}
  \label{tab:expanded_grid}
  \centering
  \small
  \begin{tabular}{lcccc}
    \toprule
    Kernel & vs.\ cuDNN & Wins & vs.\ FA4 & Wins \\
    \midrule
    MHA & $+9.97\%$ & 68/80 & $+3.03\%$ & 74/80 \\
    GQA & $+7.07\%$ & 75/80 & $+2.45\%$ & 68/80 \\
    \bottomrule
  \end{tabular}
\end{table}

Within this grid, the most demanding subset is long-context inference. Table~\ref{tab:expanded_longctx} reports the geometric mean over sequence lengths $\{65536, 131072\}$ and both attention masks at each batch size. Gains over cuDNN are substantially larger at the higher batch sizes, while gains over FA4 remain positive throughout.

\begin{table}[h]
  \caption{Long-context subset of the 80-configuration grid. Each entry is the geometric mean over sequence lengths $\{65536, 131072\}$ and both causal and non-causal masking.}
  \label{tab:expanded_longctx}
  \centering
  \small
  \begin{tabular}{llcc}
    \toprule
    Kernel & Batch & vs.\ cuDNN & vs.\ FA4 \\
    \midrule
    \multirow{4}{*}{MHA} & 2  & $+11.94\%$ & $+1.77\%$ \\
                         & 4  & $+21.10\%$ & $+1.35\%$ \\
                         & 8  & $+23.92\%$ & $+1.89\%$ \\
                         & 16 & $+22.70\%$ & $+3.42\%$ \\
    \midrule
    \multirow{4}{*}{GQA} & 2  & $+7.68\%$ & $+1.07\%$ \\
                         & 4  & $+8.59\%$ & $+1.15\%$ \\
                         & 8  & $+9.04\%$ & $+1.24\%$ \\
                         & 16 & $+9.13\%$ & $+1.29\%$ \\
    \bottomrule
  \end{tabular}
\end{table}

The speedups are therefore not an artifact of the benchmark suite that defined the search objective: they hold across two orders of magnitude in sequence length, and at batch--length combinations the scoring function never covered, since evolution tied each sequence length to a single batch size.
\end{crebutsec}

\section{Ablation Study}
\label{app:ablation}

\begin{crebutsec}
To quantify how much each element of the AVO loop contributes, we disable one component of the full system at a time and re-run the search with everything else unchanged. We remove the self-supervision mechanism of Section~\ref{sec:method_loop} and, separately, the knowledge base $\mathcal{K}$ of Section~\ref{sec:method_formal}. All conditions share the same backbone model, tools, compiler, correctness tests, benchmark harness, GPU, and 24-hour budget per task; only the ablated component differs. We evaluate on GQA decode (H100) and fused FP8 MoE (B200), following the task setup of FlashInfer-Bench~\citep{xing2026flash}.

\begin{table}[h]
  \caption{Ablation of AVO under a matched 24-hour budget per task. Each row disables one component of the full system. Lower latency is better.}
  \label{tab:ablation}
  \centering
  \small
  \begin{tabular}{lcc}
    \toprule
    Configuration & GQA decode latency & FP8 MoE latency \\
    \midrule
    Full AVO                                      & $\mathbf{18.39}~\mu s$ & $\mathbf{327.16}~\mu s$ \\
    AVO without self-supervision                  & $34.57~\mu s$          & $484.01~\mu s$ \\
    AVO without the knowledge base $\mathcal{K}$  & $18.84~\mu s$          & $373.69~\mu s$ \\
    \bottomrule
  \end{tabular}
\end{table}

Disabling self-supervision costs the most: full AVO attains $46.8\%$ lower latency on GQA decode and $32.4\%$ on FP8 MoE, even though the variation operator retains the knowledge base, the archive, and the evaluator. Across the search trajectories, the stagnation episodes that trigger self-supervision recur in three forms: repairing one compilation error only to expose another; oscillating between faster-but-incorrect and correct-but-slower candidates; and regressing throughput through increased register pressure or reduced parallelism. Sustained progress over a long horizon therefore depends on monitoring and redirecting the trajectory, not only on implementing and testing individual candidates. Removing the knowledge base costs $2.4\%$ and $12.5\%$ respectively, a consistent but substantially smaller effect.
\end{crebutsec}

\section{Toward Population-Level AVO}
\label{app:population}

\begin{crebutsec}
The main experiments instantiate AVO in a single-lineage regime (Section~\ref{sec:method_loop}). Because AVO replaces the variation operator rather than the population structure, it is compatible with both unary mutation and multi-parent recombination.

\paragraph{Lineage-internal semantic crossover.}
Committed solutions, their scores, and associated artifacts remain in the archive $\mathcal{P}_t$ and are directly accessible to the agent, which can inspect several historical kernels and integrate compatible optimizations into a single candidate. We call this \emph{lineage-internal semantic crossover}: it recombines mechanisms rather than source text, but all parents share one trajectory.

\paragraph{Cross-lineage population sharing.}
We compare eight isolated AVO runs, each with a private population, against eight lineages whose populations are synchronized every four hours, exposing each lineage to kernels and measurements found by the others. Both settings use eight four-hour windows (32-hour wall clock, up to 256 aggregate GPU-hours per workload), with model, initialization, prompts, hardware, and evaluator unchanged. Tasks are taken from FlashInfer-Bench~\citep{xing2026flash} and run on H100 GPUs.

\begin{table}[h]
  \caption{Preliminary cross-lineage study under a matched aggregate budget. Lower latency is better; the isolated baseline is the \emph{best} of eight independent replicas.}
  \label{tab:population}
  \centering
  \small
  \begin{tabular}{lccc}
    \toprule
    Evolutionary regime & GQA decode & MLA decode & MLA prefill \\
    \midrule
    Best isolated AVO run (8 replicas)   & $15.54~\mu s$ & $20.58~\mu s$ & $32.90~\mu s$ \\
    Synchronized population (8 lineages) & $\mathbf{14.65}~\mu s$ & $\mathbf{17.98}~\mu s$ & $\mathbf{31.57}~\mu s$ \\
    \midrule
    Latency reduction                    & $5.7\%$ & $12.6\%$ & $4.0\%$ \\
    \bottomrule
  \end{tabular}
\end{table}

Even against the best of eight isolated runs, simple periodic sharing reduces geomean latency by $4.0\%$ to $12.6\%$. We report this as preliminary evidence that AVO benefits from cross-lineage solutions, not as a tuned population-based system; lifting the agent itself to the population level, so that it reconciles parents carrying complementary optimizations, is a natural next step that we leave to future work.
\end{crebutsec}

\section{Limitations and Future Work}
\label{app:discussion}

\paragraph{Broader hardware coverage.}
Our \cready{main} evaluation targets server-grade NVIDIA Blackwell (B200) GPUs, where cuDNN and FlashAttention-4 have already absorbed months of expert tuning and therefore provide the most stringent test of autonomous kernel optimization. \cready{We additionally evaluate AVO on Blackwell B300 (Appendix~\ref{app:more_tasks_linear}) and Hopper H100 (Appendices~\ref{app:benchmark_evo},~\ref{app:ablation}, and~\ref{app:population}). Extending AVO to other accelerator families and to resource-constrained client and edge devices, and characterizing how the discovered optimization patterns transfer across architectures, remain future work.}

\begin{crebutsec}
\paragraph{No convergence guarantee.}
AVO provides no formal guarantee of converging to a global optimum: the trajectories in Figures~\ref{fig:evolution_causal_trajectory} and~\ref{fig:evolution_noncausal_trajectory} show diminishing returns rather than optimality, and the quality of the final kernel depends on the search budget and on the capability of the underlying model. The evolved kernels also remain open to further refinement, both by human experts and by complementary automated optimization tools.
\end{crebutsec}

\paragraph{Co-evolving the knowledge base.}
In our current setup, the knowledge base $\mathcal{K}$ is fixed throughout the evolutionary run. While this is sufficient to reach state-of-the-art performance on the benchmarked configurations, it leaves the agent reliant on a static body of external material that does not capture the lessons accumulated during the search itself. A natural extension is to co-evolve $\mathcal{K}$ alongside the candidate solutions, allowing the agent to distill profiling observations, failed optimization attempts, and recurring hardware-level patterns into the knowledge base as the run progresses.

\end{document}